\documentclass[runningheads]{llncs}

\usepackage[margin=1.23in]{geometry}
\usepackage{setspace}
\usepackage{graphicx}
\usepackage{epsf,float,epsfig}

\usepackage{amsmath}
\usepackage{amssymb}
\usepackage{natbib}

\begin{document}

\title{Unifying Decision-Making: a Review on Evolutionary Theories on Rationality and Cognitive Biases}

\author{Catarina Moreira}

\authorrunning{C. Moreira}

  \institute{   
      School of Business, University of Leicester\\
     University Road, LE1 7RH Leicester, United Kingdom\\  
         \email{cam74@le.ac.uk} 
}

\maketitle

\begin{abstract}

In this paper, we make a review on the concepts of rationality across several different fields, namely in economics, psychology and evolutionary biology and behavioural ecology. We review how processes like natural selection can help us understand the evolution of cognition and how cognitive biases might be a consequence of this natural selection. In the end we argue that humans are not irrational, but rather rationally bounded and we complement the discussion on how quantum cognitive models can contribute for the modelling and prediction of human paradoxical decisions.

\end{abstract}

\doublespace

\section{Introduction}

Rationality is one of the oldest and yet still on going research topics in the scientific community. Specially in the modern world where computational tools are used for predictive analyses of human behaviour and for the development of more sophisticated decision-making models that are able to contribute for a general theory for decision-making. Rationality is a term that has been widely debated across several fields of the literature with different discussions in Economics, Psychology and behavioural Ecology / evolutionary Biology.

In 1944, the Expected Utility theory was axiomatised by the mathematician John von Neumann and the economist Oskar Morgenstern, and became one of the most significant and predominant rational theories of decision-making~\citep{Neumann53}. The Expected Utility Hypothesis is characterised by a specific set of axioms that enable the computation of the person's preferences with regard to choices under risk~\citep{Friedman52}. By risk, we mean an uncertain event that can be measured and quantified. In other words, choices based on {\it objective probabilities}. Under this theory, human behaviour is assumed to maximise an utility function and by doing so, the person would be acting in a fully rational setting. This means that human psychological processes started to be irrelevant as long as human decision-making obeys to some set of axioms~\citep{Glimcher14}. 

 In 1953, Allais proposed an experiment that showed that human behaviour does not follow these normative rules and violates the axioms of Expected Utility, leading to the well known Allais paradox~\citep{Allais53}. Later, in 1954, the mathematician Leonard Savage proposed an extension of the Expected Utility hypothesis, giving origin to the Subjective Expected Utility~\citep{savage54}. Instead of dealing with decisions under risk, the Subjective Utility theory deals with {\it uncertainty}. Uncertainty usually described situations that involve ambiguous / unknown information. Consequently, it is specified by subjective probabilities. In 1961, Daniel Ellsberg proposed an experiment that showed that human behaviour also contradicts and violates the axioms of the  Subjective Expected Utility theory, leading to the Ellsberg paradox~\citep{Ellsberg61}. In the end, the Ellsberg and Allais paradoxes show that human behaviour does not follow a normative theory and, consequently, tends to violate the axioms of rational decision theories. In summary, when dealing with preferences under uncertainty, it seems that models based on normative theories of rational choice tend to tell how individuals {\it must} choose, instead of telling how they {\it actually} choose~\citep{Machina09}. 

The separation of economics from psychology made these two research fields take their own separate paths in terms of human decision-making. In one extreme, economics was built up from a set of strong normative assumptions that assume that to be fully rational means to obey to some set of axioms. Everything that is not chosen according to these normative axioms lead to irrational behaviour. 

In this paper, we will put together some major scientific contributions from different fields that could help to unify interdisciplinary knowledge towards a more unified decision model.

\section{The Different Types of Rationality}

\begin{figure}[h!]
\resizebox{\columnwidth}{!} {
\includegraphics{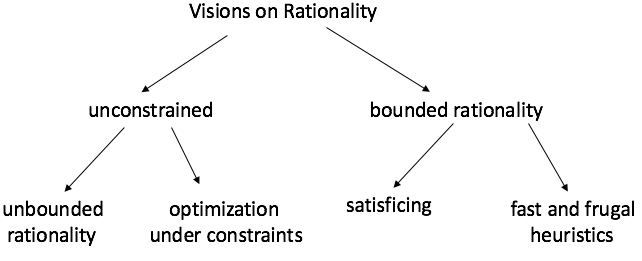}
}
\caption{Different types of rationality accross different fields as suggested in the work of~\citet{Gigerenzer01}.}
\label{fig:rationality}
\end{figure}

The psychologist Gerd Gigerenzer makes a distinction between the degrees of rationality that a theory of decision-making should incorporate~\citep{Gigerenzer01}. Disciplines like Economics, Cognitive Science, Biology, etc, assume that both humans and animals have unlimited information, unlimited computational power and unlimited time to make a decision. And consequently, as long as these decisions follow the axioms of the expected utility theory, then they are optimal and fully rational. This kind of rationality is called {\it unbounded rationality}. Figure~\ref{fig:rationality} shows the different kinds of rationality that one can find across different disciplines as proposed by~\citet{Gigerenzer01}. Figure~\ref{fig:rationality}, makes a distinction between two unconstrained theories of rationality: the unbounded rationality and the optimisation under constraints. The difference is that the later has a stopping rule, which measures and stops at the point where the cost for further search exceeds the benefits of the decision.

Although unbounded rationality is widely used across several fields (specially in Economics), it is not a concept that can be applied in real world decision scenarios. Both humans and animals have limited resources: they need to make inferences under uncertain aspects of the world with limited knowledge, limited computational power and limited time. For instance, an animal that is exposed in the wild looking for food needs to make the decision if it will either remain looking for food being completely exposed to his predators or if he should hide. If he has unlimited time to make this decision, this could evolve the computation of several alternatives in order to choose the one that grants the animal the highest utility. However, if the animal looses too much time in this reasoning process, he could get attacked by its predators. So, there are constraints on resources when making decisions under uncertainty. One can even argue that both humans and animals are fully rational under these constraints: we try to find the best option with the limited time, computational power and information that we have. This is usually called {\it bounded rationality} and usually incorporates heuristics (or rules of thumb) which are gathered by experience and memory in order to turn the decision process faster. Bounded rationality was a term introduced by~\citet{Simon55} where he suggested that decision-makers are seen as  satisficers, who seek a solution that is satisfiable under constraints in time, information and computation, rather than the optimal one. 

\section{Biological System: Does Natural Selection Produce Rational Behaviour?}

In evolutionary Biology, the {\it fitness} of an individual is a measure how good an individual (or a species) is at leaving offsprings in the next generation. This is a very important concept, since natural selection consists on the heritable variations that a trait can suffer and consequently influence the fitness level of an individual~\citep{Stevens08}. The premise is that, the more surviving descendants an individual produces, the higher is the fitness level.

Since natural selection is the process by which biological evolution occurs, leading to a species with more fitness, many scholars have posed the question whether natural selection has any role in rational behaviour~\citep{Stevens08,Houston07,Santos15}.

Like humans, animals also need to make decisions that grants them higher profits. In other words, they need to take into account temporal delays and uncertainty as well as potential payoffs in pursuing different actions~\citep{Santos15}. In this sense, human decisions under economic scenarios can have some analogies with the type of decisions that animals are faced when they are foraging for food or seeking mates: in both scenarios it is assumed that they prefer choices that grant them higher profits / food / fitness levels. The question that arises in this context is if, under an evolutionary point of view, natural selection leads to choices that are in accordance with the expected utility hypothesis, then how did human biases emerge?

One of the cognitive biases pointed in~\citet{Tversky86} work is the {\it framing effect}. In framing effects, people react differently to some choice depending on how that choice is presented. For instance, people tend to avoid risk when a positive frame is presented but seek risks when a negative frame is presented~\citep{Kahneman82book}. Research shows that these framing effects are not singular to humans, but they also occur in primates. For instance, \citet{Chen06} made an experiment where monkeys could trade tokens with human experimenters in exchange for food. In the end, the amount of food that each monkey received form the human experimenters was the same. The difference was that one experimenter showed the monkey one piece of apple and then added an extra one (the gains experimenter), while the other showed two pieces of apple to the monkey, but removed one (the losses experimenter). Although the monkeys received the same payoff, they preferred the gains experimenter over the losses one, indicating that monkeys also fall under the framing effect like humans~\citep{Santos15}.

Another study with monkeys conducted by~\citep{Lakshminarayanan11} showed that monkeys also revealed a reflection effect. The reflection effect explains that we have opposite 'risk preferences' for uncertain choices, depending on whether the outcomes is a possible gain or a loss. In the study of~\citep{Lakshminarayanan11}, monkeys  tended to seek out more risk when dealing with losses when compared to gains. 

These works suggest that cognitive biases are not singular to humans, but they occur in non-human beings as well. These findings could imply some ancestral roots in our cognitive system. Returning to the question of whether natural selection plays a role in a fully rational cognitive system, studies suggest that it is not the case~\citep{Stevens08}. Natural selection does play a role in optimising the fitness of an individual, like it is stated in the work of~\citep{Santos15}. If the environment is constant, then the process of natural selection of a species reaches its optimal fitness with time. It seems that context does play an important role in natural selection. 

Following the discussion on~\citep{Santos15}, in evolutionary biology it is agreed that it may be rational for an individual to make biased and inconsistent preferences, if these preferences point towards the maximisation of the fitness~\citep{Kacelnik06}. In other words, individuals my be acting rational by falling into cognitive biases that may lead them to the survival of the species. This can be connected with the notion of bounded rationality that it was presented in the previous section. These cognitive biases produce rules of thumb for individuals that, although they are not optimal, they lead to the best possible decision given the limited constraints on time, computational power and information. In other words, they are able to reach a decision that grants them a high return (not the most optimal one), using less computational resources through the usage of heuristics. Note that, the notion of heuristic consists precisely in a {\it shortcut} that usually leads to the desired outcome, but sometimes and get lead to wrong decisions and outcomes~\citep{Shah08}. 

Experiments conducted on birds also show the importance of context in behavioural ecology. Studies have shown that both birds and insects deviate significantly from their choices when the context varies~\citep{Kacelnik02}. Following the arguments in~\citep{Santos15}, in behavioural ecology, the energetic increase of an individuals fitness level is non-linear. This means that a single unit of food has a significant impact on an individual that has a low level of energy (an individual that is hungry), but it would have a lower impact (or even a negative impact) on an individual that is already in a high energy state (not hungry). So, the risk preferences that both humans and animals choose in different contexts might be optimising the fitness measure under a biological point of view depending on the context where they are. 

Summarising, natural selection alone does not lead to optimal rational and cognitive behaviour, but it always leads to the optimisation of the fitness measure. If the context (or environment) where the individual is contained is constant, then natural selection can indeed lead to fully rational and optimal choices. However, these are exceptional scenarios. Throughout time, environments keep changing and individuals are required to adapt in order to optimise their fitness measure and to survive. Under a constant change of context and environment, individuals might be more susceptible to fall under cognitive bias and apply rules of thumb and heuristics that can help them get the decision outcome taking into considerations the limitations in knowledge regarding the new environmental context, limitations in computational power and limitations in time. Ultimately, individuals are not being irrational, but rather rational given all these constraints. In economics, cognition and other disciplines, this behaviour is seen as purely irrational, since it is violating the axioms of expected utility theory.

The question that now arises is whether there is a mathematical theory that is suitable to mode these cognitive biases and violations to the axioms of expected utility in a more general and elegant way that could be used across different fields that analyse cognition and decision-making through different perspectives. We suggest that the answer to this question is positive and that the answer can be pointed towards the mathematical formalisms of quantum mechanics.

\section{Quantum Cognition}

Motivated by cognitive biases identified by Tversky and Kahnman, researchers started to look for alternative mathematical representations in order to accommodate these violations~\citep{Tversky74,Tversky86,Kahneman72,Tversky92}. Although in the 40's, Niels Bohr had defended and was convinced that the general notions of quantum mechanics could be applied in fields outside of physics~\citep{Murdoch89}, it was not until the 1990's that researchers started to actually apply the formalisms of quantum mechanics to problems concerned with social sciences. It was the pioneering work of~\citet{Aerts94} that gave rise to the field {\it Quantum Cognition}. In their work,~\citet{Aerts94} designed a quantum machine that was able to represent the evolution from a quantum structure to a classical one, depending on the degree of knowledge regarding the decision scenario. The authors also made several experiments to test the variation of probabilities when posing {\it yes / no} questions. According to the authors, the experiment suggested that when participants did not have a predefined answer regarding a question, then the answer would be updated at the moment the question was asked, indicating that the answer is highly contextual. In other words, the answer is formed by the interaction between the participant and the person asking the question. This  deviates from classical statistics, because the answer is not dependent on the participant's beliefs. A further discussion about this study can be found in the works of~\citep{Aerts95quantum_structures,Aerts96,Aerts98,Gabora02,Aerts11}. 

Quantum cognition has emerged as a research field that aims to build cognitive models using the mathematical principles of quantum mechanics. Given that classical probability theory is very rigid in the sense that it poses many constraints and assumptions (single trajectory principle, obeys set theory, etc.), it becomes too limited (or even impossible) to provide simple models that can capture human judgments and decisions since people are constantly violating the laws of logic and probability theory~\citep{Busemeyer15,Busemeyer14,Aerts14}. 

Following the lines of thought of~\citet{Sloman14}, people have to deal with missing / unknown information. This lack of information can be translated into the feelings of ambiguity, uncertainty, vagueness, risk, ignorance, etc~\citep{Zadeh06}, and each of them may require different mathematical approaches to build adequate cognitive / decision problems. Quantum probability theory can be seen as an alternative mathematical approach to model such cognitive phenomena. 

\begin{figure}
\resizebox{\columnwidth}{!} {
\includegraphics{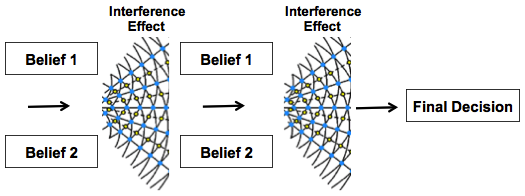}
}
\caption{Example of quantum superposition and quantum interference effects. Human beliefs are seen as indefinite states that are occuring in the human mind at the same time. If we model these beliefs as waves that propagate, then tehy can interfere with each other thourgh quantum interference effects leading to different decision outcomes.  }
\label{fig:inter}
\end{figure}

The heart of quantum cognition is to use concepts of quantum mechanics such as superposition and quantum interference effects in order to accommodate the paradoxical findings found in the literature. For instance, in quantum information processing, information is modelled via wave functions and therefore they cannot be in definite states. Instead, they are in an indefinite quantum state called the $superposition$ state. That is, all beliefs are occurring on the human mind at the same time, instead of the classical approach which considers that each belief occurs in each time frame. According to cognitive scientists, this effect is responsible for making people experience uncertainties, ambiguities or even confusion before making a decision. At each moment, one belief can be more favoured than another, but all beliefs are available at the same time. In this sense, quantum theory enables the modelling of the cognitive system as if it was a wave moving across time over a state space until a final decision is made. From this superposition state, uncertainty can produce different waves coming from opposite directions that can crash into each other, causing an interference distribution. This phenomena called {\it quantum interference effects} is the heart of quantum cognition and it can never be obtained in a classical setting. When the final decision is made, then there is no more uncertainty. The wave collapses into a definite state. Thus, quantum information processing deals with both definite and indefinite states~\citep{Busemeyer12book}. Figure~\ref{fig:inter} shows an example of quantum superposition and quantum interference effects. 

Some researchers argue that quantum-like models do not offer many underlying aspects of human cognition (like perception, reasoning, etc). They are merely mathematical models used to fit data and for this reason they are able to accommodate many paradoxical findings~\citep{Lee13}. Indeed quantum-like models provide a more general probability theory that use quantum interference effects to model decision scenarios, however they are also consistent with other psychological phenomena (for instance, order effects)~\citep{Sloman14}. In the book of~\citet{Busemeyer12book}, for instance, the feeling of uncertainty or ambiguity can be associated to quantum superpositions, in which assumes that all beliefs of a person occur simultaneously, instead of the classical approach which considers that each belief occurs in each time frame. 

The non-commutative nature of quantum probability theory enables the exploration of methods capable of explaining violations in order of effects. Order of effects is a fallacy that consists in querying a person in one order and then posing the same questions in reverse order. Through classical probability theory, it would be expected that a person would give the same answers independently of the order of the questions. However, empirical findings show that this is not the case and that people are influence by the context of the previous questions. Moreover, the existence of quantum interference effects also enables the exploration of models that accommodate other typed of paradoxical findings such as disjunction and conjunction errors~\citep{Tversky92,Tversky83Uncertainty}. In summary, quantum probability theory is a general framework, which can naturally explain various decision-making paradoxes without falling into the restrictions of classical probability theory. 

There are many quantum-like models proposed in the literature~\citep{Aerts13concepts,Aerts13,Busemeyer09,busemeyer06,Khrennikov10contextual,Yukalov11}. For the purposes of this paper, we will present a model that is based on modularity. Like it was presented, both humans and non-humans have limited computation and information processing capabilities. In order to reason about something, one needs to combine small pieces of information in order to make a decision about it~\citep{Griffiths08}. One model that can represent this modularity is the Quantum-Like Bayesian Network originally proposed by~\citet{Moreira14, Moreira16}.

\section{Modularity and Quantum-Like Bayesian Networks for Decision-Making}

Bayesian Networks are one of the most powerful structures known by the Computer Science community for deriving probabilistic inferences (for instance, in medical diagnosis, spam filtering, image segmentation, etc)~\citep{koller09prob}. They provide a link between probability theory and graph theory. And a fundamental property of graph theory is its modularity: one can build a complex system by combining smaller and simpler parts. It is easier for a person to combine pieces of evidence and to reason about them, instead of calculating all possible events and their respective beliefs~\citep{Griffiths08}. In the same way, Bayesian Networks represent the decision problem in small modules that can be combined to perform inferences. Only the probabilities which are actually needed to perform the inferences are computed.

This process can resemble human cognition~\citep{Griffiths08}. While reasoning, humans cannot process all possible information, because of their limited capacity. Consequently, they combine several smaller pieces of evidence in order to reach a final decision.

\subsection{Bayesian Networks}\label{sec:bn}

\begin{figure}
\resizebox{\columnwidth}{!} {
\includegraphics{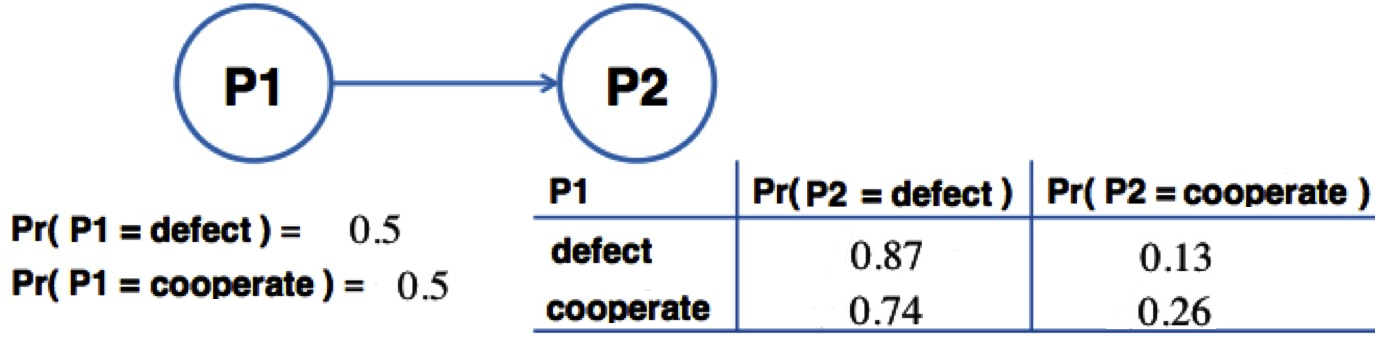}
}
\caption{Example of classical Bayesian network with two random variables (nodes). }
\label{fig:bn}
\end{figure}

A classical Bayesian Network can be defined by a directed acyclic graph structure in which each node represents a different random variable from a specific domain and each edge represents a direct influence from the source node to the target node. The graph represents independence relationships between variables and each node is associated with a conditional probability table which specifies a distribution over the values of a node given each possible joint assignment of values of its parents. This idea of a node, depending directly from its parent nodes, is the core of Bayesian Networks. Once the values of the parents are known, no information relating directly or indirectly to its parents or other ancestors can influence the beliefs about it~\citep{koller09prob}. Figure~\ref{fig:bn}, shows the representation of a Bayesian Network.

Bayesian Networks are based on probabilities. Probability theory is a formal framework that is capable of representing multiple outcomes and their likelihoods under uncertainty. Uncertainty is a consequence of various factors: limitations in our ability to observe the world, limitations in our ability to model it and possibly even because of innate nondeterminism~\citep{koller09prob}. If we take into account the basic structures behind probabilistic systems, we will find that in one extreme there is the full joint probability distribution and in another extreme there is the Na\"{i}ve Bayes model.

The full joint probability distribution corresponds to the entire knowledge of a system. That is, it represents the probabilities of all possible atomic events in a domain. However, this explicit representation of probability distributions is a highly demanding process and most of the times such information is not possible to obtain as a full. Even in its simplest case, when $N$ variables of a distribution contain binary values, one would need to compute $2^N$ entries. Under a computational point of view, the manipulation of this distribution is a heavy and expensive process and the probability distribution tables become too large to be stored in memory.

In the other extreme, there is the Na\"{i}ve Bayes model. Reasoning about any realistic domain always requires that some simplifications are made. The very act of preparing knowledge to support reasoning requires that we leave many facts unknown, unsaid or crudely summarised. The Na\"{i}ve Bayes model assumes that all random variables are independent of each other given a class. This strong independence assumption greatly reduces the computational costs when compared to full joint probability distribution. However, the independence assumption rarely holds when it comes to modelling real world events, leading to more inaccurate models.

Bayesian Networks can be seen as a framework that falls between the full joint probability distribution and the Na\"{i}ve Bayes Model. It is a compact representation of high dimensional probability distributions by combining conditional parameterisations with conditional independences in graph structures~\citep{koller09prob}. 

A Bayesian Network can be understood as the representation of a full joint probability distribution through conditional independence statements. This way, a Bayesian Network can be used to answer any query about the domain by combining (adding) all relevant entries from the joint probability.

\subsection{Quantum-Like Bayesian Networks}

In the work of \citep{Moreira14,Moreira16}, the authors suggest to define the Quantum-Like Bayesian Network by replacing real probability numbers by quantum probability amplitudes, which are complex numbers. 

\begin{figure}[h!]
\centering
\includegraphics[scale=0.5]{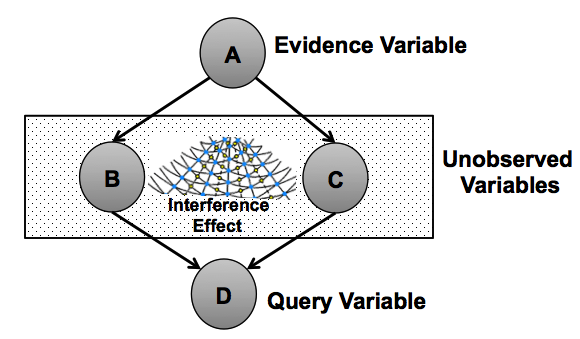}
\caption{Example of a quantum-like Bayesian network producing quantum interference effects when random variables $B$ and $C$ are not observed.}
\label{fig:qbn}
\end{figure}

A complex number is a number that can be expressed in the form $z = a + ib$, where $a$ and $b$ are real numbers and $i$ corresponds to the imaginary part, such that $i^2 = -1$. Alternatively, a complex number can be described in the form $z = \left| r \right| e^{i\theta}$, where $|r| = \sqrt{a^2 + b^2}$. The $e^{i\theta}$ term is defined as the phase of the amplitude and corresponds to the angle between the point expressed by $(a,b)$ and the origin of the plane. These amplitudes are related to classical probability by taking the squared magnitude of these amplitudes through Born's rule.

The general idea of a Quantum-Like Bayesian network is that when performing probabilistic inference, the probability amplitude of each assignment of the network is propagated and influences the probabilities of the remaining nodes. In other words, \emph{every} assignment of \emph{every} node of the network is propagated until the node representing the query variable is reached. Note that, by taking multiple assignments and paths at the same time, these trails influence each other producing interference effects. If a node (or a random variable) is not observed, then it remains in a superposition state and it can create quantum interference effects that can be used to accommodate the several paradoxical findings in the literature. Figure~\ref{fig:qbn} shows the underlying idea of the quantum-like Bayesian Network.

\begin{figure}[h!]
\resizebox{\columnwidth}{!} {
\includegraphics[scale=0.4]{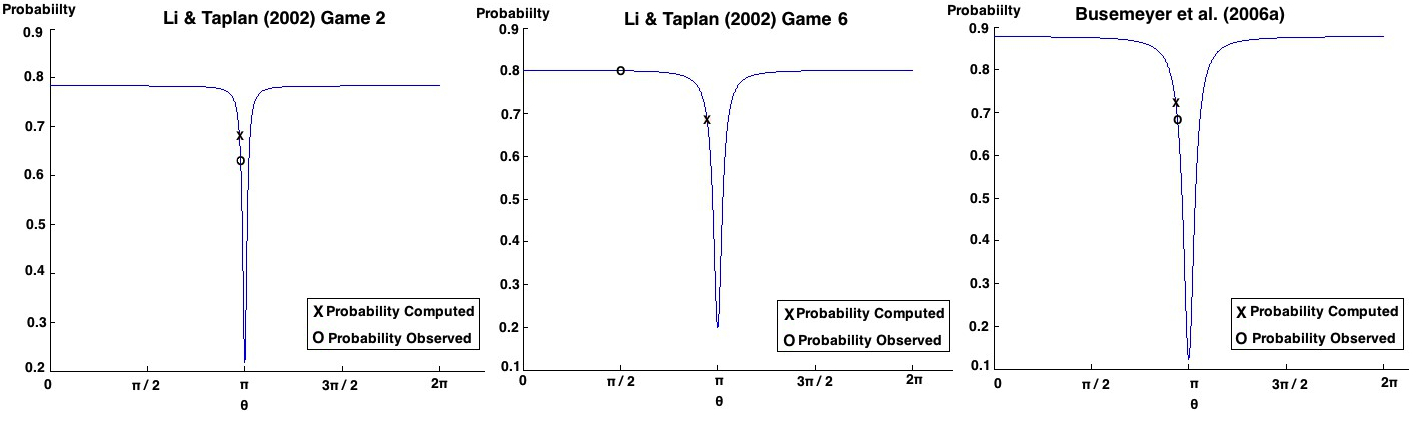}
}
\caption{Example of all possible quantum interference effects under the Prisoner's Dilemma experiments made by~\citep{Li02} and~\citep{Busemeyer06proceed}. In the figures, it is shown the estimation of the quantum interference terms (\it{Probability Computed}) thourgh an heuristic function proposed in~\citep{Moreira16} and the probability outcome of the participants observed in the experiments ({\it Probability Observed}).}
\label{fig:interference}
\end{figure}

Recent research shows that quantum interference effects can serve as a fitting function that enables an extra parametric fitting layer when compared with the classical model. Very generally, a probabilistic inference on a quantum-like Bayesian network with two nodes $A$ and $B$, expressed as probability amplitudes as $\psi_A$ and $\psi_B = $ respectively, is given by
\[ Pr( B ) = | \psi_A |^2 + | \psi_{B } |^2 + 2 \psi_A \psi_{B } \cos \left( \theta_A - \theta_B \right), \]
where $\theta_A$ and $\theta_B$ are the quantum interference parameters. Figure~\ref{fig:interference} shows all possible quantum interference effects that can emerge from the Prisoner's Dilemma experiment proposed by~\citet{Li02} and~\citet{Busemeyer06proceed} in order to identify disjunction errors. Note that a disjunction error corresponds to the judgment of two events to be least as likely as either of those events~\citep{Carlson89}. This way, by choosing the right interference term, one can explain the human cognitive bias and even predict human {\it rationally bounded} decisions. 

\section{Conclusion}

In this work, we made a review of how different fields of the literature deal with the concept of rationality. 

Since the axiomatization of the expected utility theory, it has been assumed that humans make optimal decisions, always choosing the preference or the decision that leads to higher gains or higher profits. Every decision that does not maximise this utility is considered irrational and not optimal. 

Studies from the last decades show that humans constantly violate the axioms of expected utility theory and now, evolutionary biology and ecological studies also demonstrate that humans are not alone is this subject: both humans and non-humans systematically deviate from what rational choice predicts. If the cognitive bias is shared between humans and non-humans, then it means that we share an ancient and primordial set of heuristics and rules of thumb that helped us in the surviving process of the species.

From the evolutionary Biology perspective, natural selection plays a role in finding the optimal fitness measure. By fitness, we mean the capacity of an individual reproducing and surviving. If an individual remains in a constant environment, then the cognitive abilities will be optimised and indeed they can lead to fully optimal and rational decisions. But these scenarios are scarce. Environment is constantly changing and individuals need to adapt to this change. Given that individuals have limited processing power and limited time, they need to take actions with the limited information that they have. They can use cognitive biases to speed up these decisions. Although they are not fully optimal, they are the best that they can take given the constraints of time, knowledge and computational power. So, under an evolutionary point of view, individuals are rationally bounded, while in disciplines such as economics or cognitive science, these types of decisions are seen irrational.

We ended this paper by introducing quantum mechanics as a general mathematical approach that could take into account quantum superpositions and quantum interference effects to model, explain and predict cognitive biases. We presented a network based model that has been widely studied in the literature that combines small pieces of information in order to perform more complex inferences. 

To conclude, there are many disciplines that have large contributions over the topic of rationality and how cognitive biases can be explained through different perspectives. If one wants to model a unified theory of decision-making, one needs to start having a holistic view of the subject in order to produce more robust and more predictive models for decision-making.

 \bibliographystyle{frontiersinSCNS_ENG_HUMS}

\end{document}